\documentclass[runningheads,a4paper]{llncs}

\usepackage[cmex10]{amsmath}
\usepackage{amssymb}
\usepackage{llncsdoc}
\usepackage{mathptmx}       
\usepackage{helvet}         
\usepackage{courier}        
\usepackage{makeidx}         
\usepackage[pdftex]{graphicx}
\usepackage{epstopdf}
\DeclareGraphicsExtensions{.pdf,.jpeg,.png,.eps}
\usepackage{multirow}        
\usepackage[bottom]{footmisc}
\usepackage{url}
\usepackage[linesnumbered,ruled,vlined]{algorithm2e}
\usepackage{array}
\usepackage{booktabs}
\usepackage[caption=false,font=footnotesize]{subfig}
\usepackage{xcolor,colortbl}
\usepackage{bm}
\usepackage{subfig}
\usepackage{geometry}
\usepackage{graphicx}
\usepackage{booktabs}
\usepackage{float}

\DeclareMathSymbol{\R}{\mathalpha}{AMSb}{"52}

\definecolor{Gray}{gray}{0.85}
\definecolor{LightCyan}{rgb}{0.88,1,1}

\begin{document}
\mainmatter 

\title{Enhancing High-Energy Particle Physics Collision Analysis through Graph Data Attribution Techniques}

\author{A. Verdone$^1$, A. Devoto$^1$, C. Sebastiani$^{4}$, J. Carmignani$^4$, M. D'Onofrio$^4$, S. Giagu$^{2,3}$, S. Scardapane$^{1,3}$ and M. Panella$^1$}

\titlerunning{Enhancing High-Energy Particle Physics Collision Analysis through Graph Data Attribution Techniques}
\authorrunning{Verdone et al.}
\institute{$^1$ Department of Information Engineering, Electronics and Telecommunications (DIET), ``Sapienza'' University of Rome, Via Eudossiana 18, 00184, Rome, Italy. \\
$^2$ Department of Physics, ``Sapienza'' University of Rome, Piazzale A. Moro 5, 00185, Rome, Italy. \\
$^3$ INFN Sezione di Roma, Piazzale Aldo Moro, 5, Rome, 00185, Italy, UE. \\
$^4$ Department of Physics, University of Liverpool, Oxford Street Liverpool, L69 7ZE, United Kingdom.\\
Email: \{alessio.verdone; alessio.devoto; stefano.giagu; simone.scardapane; massimo.panella\}@uniroma1.it;  \{cristiano.sebastiani; joseph.carmignani; Monica.D'Onofrio\}@cern.ch}

\maketitle

\begin{abstract}
The experiments at the Large Hadron Collider at CERN generate vast amounts of complex data from high-energy particle collisions. This data presents significant challenges due to its volume and complex reconstruction, necessitating the use of advanced analysis techniques for analysis. Recent advancements in deep learning, particularly Graph Neural Networks, have shown promising results in addressing the challenges but remain computationally expensive. The study presented in this paper uses a simulated particle collision dataset to integrate influence analysis inside the graph classification pipeline aiming at improving the accuracy and efficiency of collision event prediction tasks. By using a Graph Neural Network for initial training, we applied a gradient-based data influence method to identify influential training samples and then we refined the dataset by removing non-contributory elements: the model trained on this new reduced dataset can achieve good performances at a reduced computational cost. The method is completely agnostic to the specific influence method: different influence modalities can be easily integrated into our methodology. Moreover, by analyzing the discarded elements we can provide further insights about the event classification task. The novelty of integrating data attribution techniques together with Graph Neural Networks in high-energy physics tasks can offer a robust solution for managing large-scale data problems, capturing critical patterns, and maximizing accuracy across several high-data demand domains. 

\end{abstract}

\section{Introduction}
\label{sec:introduction}
The Large Hadron Collider (LHC) at CERN provides high-energy particle beams for experiments like ATLAS~\cite{ATLAS_2008xda}, which generate vast amounts of data from collisions. These collisions produce a vast array of particles that are detected by sophisticated experimental apparatus. The data collected are of extreme importance for understanding the fundamental nature of matter and the universe. However, the sheer scale and complexity of the data pose significant challenges for efficient and accurate analysis~\cite{Bird:2011zz}.
For example, the output from ATLAS event reconstruction can generate a data stream of more than 3.5 terabytes per second~\cite{ATLAS:2024vdo}: this enormous amount of data is then processed and analyzed by teams of scientists and researchers who use a variety of techniques and algorithms to extract meaningful information from the data. The complexity of the data is further exacerbated by the presence of missing values, outliers, and noisy data points, which can lead to inaccurate and biased results  \cite{dillon2023normalized,buss2023s,wozniak2023quantum}. The analysis of the data requires a deep understanding of the underlying physics and the ability to identify patterns and relationships that may not be immediately apparent, hence necessitating a high degree of expertise and specialized knowledge.
In recent years, machine learning and deep learning techniques have shown very promising results in addressing the challenges posed by the LHC data \cite{guest2018deep}. These methods have been successfully applied to a range of tasks, including particle identification \cite{ypsilantis1995particle}, event reconstruction \cite{staszewski2009transport}, and background subtraction \cite{crochet2002investigation}. However, the complexity and scale of the data require the development of more sophisticated and scalable methods that can effectively handle the data volume generated by the LHC experiments.
By representing the data as a graph, where particles and their interactions are denoted by nodes and edges respectively, Graph Neural Networks (GNNs) can learn high-level representations of the data that capture the complex relationships between particles and their correlations \cite{thais2022graph,ju2020graph}. This enables more accurate and efficient analyses, as well as the ability to identify patterns and relationships that may not be apparent through traditional methods. Although GNNs, such as any deep learning model, are beneficial for the analysis of large datasets, if excessively large, the computational time and efficiency of these models become costly and possibly prohibitive.
Data attribution methods have emerged as crucial tools in machine learning and data analysis scenarios \cite{dai2023training,hammoudeh2024training,nohyun2022data,k2021revisiting} to resolve these challenges by offering insights into the inner workings of complex models and shedding light on the factors that drive the predictions at the sample level. These methods provide a mean to understand the importance (or influence) of data points in driving the output of a model, thereby enhancing interpretability and trustworthiness. Data attribution methods trace a model behavior back to its training dataset, offering an effective approach to better understanding ``black-box'' neural networks. Several methods for the detection of the influence of training data have been proposed in the last years like \emph{Trak} \cite{park2023trak}, \emph{SimFluence} \cite{guu2023simfluence} or \emph{Datamodels} \cite{ilyas2022datamodels}. One of the most important methods is \emph{TracIn} \cite{pruthi2020estimating}; it utilizes loss gradients to generate relationship scores of influences between training and testing samples. It can be used also to generate influence scores between the elements of the training set, which is a useful feature for discovering anomalies inside the training dataset. 
Recently, several works have applied successfully influence analyses on large-scale generative AI tasks, such as diffusion models and large language modells (LLMs) \cite{xie2024data,kwon2023datainf,wang2023evaluating,zheng2023intriguing,georgiev2023journey}. The considered scenarios are all characterized by a huge amount of data. For example, in image classification tasks they can identify most confounding images in the training set, (e.g., multiple classes of objects inside a single image) or the presence of wrong labels inside the datasets. Removing harmful or redundant elements enhances overall performance, improving both classification metrics and reducing computational costs. 

\begin{figure}[!t]
    \centering
    \includegraphics[width=0.85\textwidth]{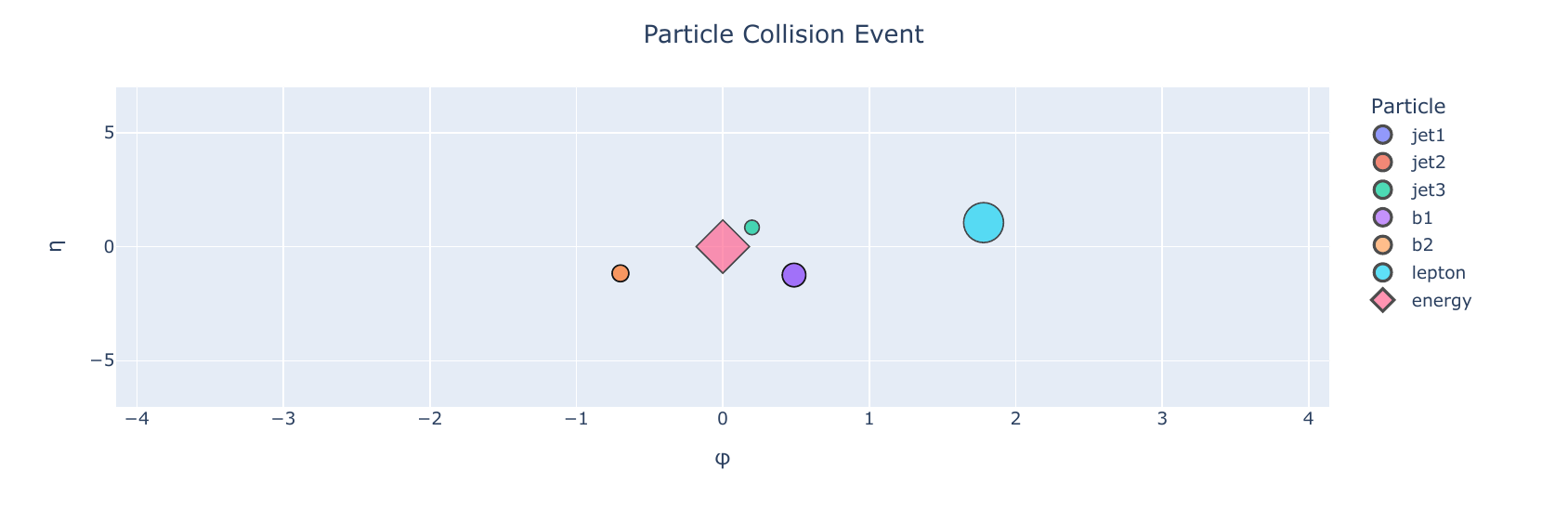}\label{fig:sub1}
		\vspace{-18pt}
    \caption{Event collision represented in a 2D plane with $\varphi$ and $\eta$ as axis. The $\varphi$-$\eta$ plane in an LHC experiment is a coordinate system used to describe the angular distribution of particles, where $\eta$ measures the particle's angle relative to the beam axis and $\varphi$ represents the azimuthal angle around the beam axis. Edges of fully connected graph are not shown for clarity.}
    \label{fig:event_collision2d}
\end{figure}

To the best of our knowledge, existing literature lacks methods that effectively combine training data attribution techniques with graph data and graph neural networks. Furthermore, the domain of high-energy physics provides a robust testing ground for evaluating our approach to tackling complex real-world problems.
Carefully selecting elements for the training set not only facilitates the management of large-scale data, which is a common challenge in physics and numerous other fields but also enables the capture of crucial patterns and relationships within the data. This approach maximizes the predictive accuracy and generalization capability across diverse domains and applications.

In this study, we integrated efficient influence analysis into the graph classification process to enhance the accuracy and efficiency of predictions regarding the collision events of high-energy particles. The pipeline can be summarized in three steps. Initially, a GNN model is used in the first training stage to classify event collision types. Then, using training checkpoints, TracIn identifies the influence relationships between samples, allowing us to discover which training samples contribute positively or negatively to the collision classification task. Finally, we can remove or replace the training dataset elements that do not improve the task and train a GNN on a selected and reduced dataset for the same task. Our experiments were conducted using a vast and extensive dataset of simulated particle collisions. This approach also enables us to perform an explainability analysis of the problem. By comprehending the characteristics of the discarded elements, as well as those defined as significant for the problem, we gain insight into both the prediction model and the subsequent downstream task. Our contribution can be resumed as: 
\begin{itemize}
\item \textbf{Integration of Influence Analysis in Classification}: We incorporated TracIn, a data attribution method, into the classification process of particle collision events using GNNs. This allowed us to identify and refine the training dataset by removing non-contributory samples, enhancing the efficiency and accuracy of the classification task.
\item \textbf{Improved Performance and Reduced Computational Costs}: By refining the dataset and focusing on significant training samples, we improved overall classification performance and reduced computational costs. This approach led to better utilization of resources and more accurate predictions.
\item \textbf{Enhanced Explainability and Insights}: Our method provides a detailed explainability analysis, offering insights into the characteristics of influential data elements. This not only helps in understanding the prediction model better but also aids in managing large-scale data by capturing critical patterns and relationships within the data.
\end{itemize}

\begin{figure}[t]
    \centering
    \includegraphics[width=0.6\textwidth]{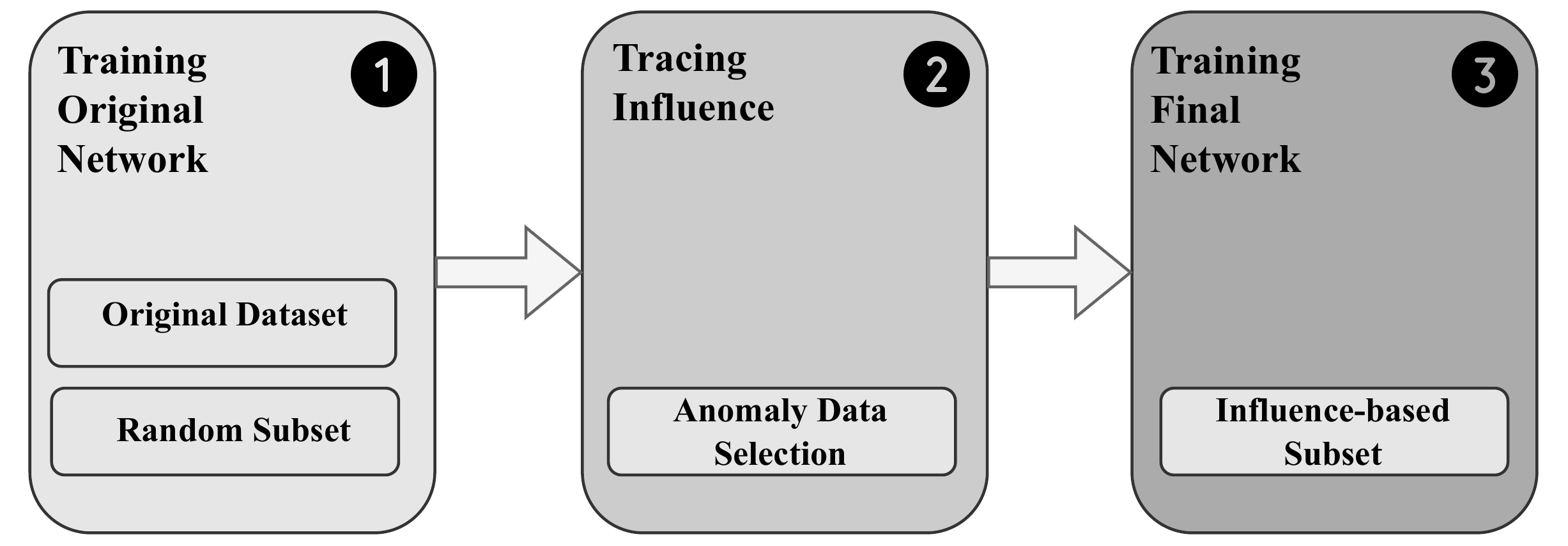}
		\vspace{-3pt}
    \caption{Our proposed methodology: we initially train the GNN network on the original full-size dataset or a subset of it. Then, we employ the saved checkpoints to compute influence values on training data: values with a higher score will be filtered out. We obtain a distilled dataset on which we perform the final training.}
    \label{fig:etichetta}
\end{figure}

\section{Related works}
\label{sec:related}
\subsection{Graph Neural Networks}
GNNs are a type of neural network designed to work with graph-structured data, able to learn representations that capture complex relationships between nodes. They are mathematical models that can be easily adapted to different tasks in the domain of the graphs, like node or graph classification and regression, edge prediction, graph generation or node clustering. They found several applications in real scenarios, such as in community detection, social network analysis, molecular property prediction or knowledge graph generation; moreover, they have found a wide scope for application in the field of particle physics \cite{thais2022graph} and high-energy physics (HEP) (e.g., particle tracking and reconstruction \cite{Duarte_HEP,ju2020graph}). The physics tasks of the LHC present many potential applications where graph neural networks have been successfully applied \cite{dezoort2023graph}. \cite{Elabd_2022} employed a GNN for the determination of charged particle trajectories in collisions. \cite{martinez2019pileup} tackled the pileup mitigation problem, the presence of parasitic low-transverse-momentum collisions, by employing a three-layer Gated Graph Neural Networks with residual connections. More recently, \cite{chatterjee2024rotation} proposed a rotation-equivariant, with respect to rotations around the jet axis, GNN  to extract novel phenomena in the standard model effective field theory (SMEFT) context from LHC collision data

\subsection{Data attribution}
Training Data Attribution (TDA) methods aim to understand the influence or importance of individual training data points on the predictions made by a machine learning model, identifying data points' influence on the model's output. Influence estimation approaches can be divided into two main classes: retraining-based and gradient-based \cite{hammoudeh2024training}. Retraining-based methods assess the influence of training data by repeatedly retraining the model using different subsets of the training set, while gradient-based influence estimators determine influence by analyzing the alignment of training and test instance gradients, either throughout the training process or at its conclusion. Retraining-based methods comprehend the simplest and more computationally expensive leave-one-out (LOO) \cite{weisberg1982residuals} or downsampling \cite{feldman2020neural}. 
More interestingly are Gradient-based methods: they typically provide closed-form TDA scores by employing gradients in an efficient and scalable way. \cite{koh2020understanding} was one of the first works in this field, by approximating the real influence effect of a training point by employing the gradients of the loss functions. TracIn \cite{pruthi2020estimating} traces loss changes on test points during the training process, while TRAK \cite{park2023trak} uses the neural tangent kernel with random projection to assess influence. These gradient-based methods have significantly reduced computational costs compared to retraining-based methods. However, they typically rely on the assumption of a first-order approximation of the loss, which can lead to performance degradation on neural networks \cite{basu2020influence,bae2022if} and be more sensitive to randomness associated with model weight initialization and training mechanisms \cite{k2021revisiting}.
The latest approach in the TDA scenario demonstrated the effectiveness of ensembling in improving TDA scores with gradient-based methods to solve these typical issues \cite{deng2024efficient,dai2023training}. Ensembling usually involves applying the TDA method to many independently trained models (e.g., averaging the final TDA scores or aggregating some intermediate terms for score calculation). Despite their effectiveness, these ensembling methods require a substantial number of ensembles to perform well, a constraint that requires an important computational cost.

\subsection{Data distillation}
Data distillation refers to the process of carefully choosing which data points to include in the training set for a deep learning model, as the quality and distribution of the training data can significantly impact the model's performance or computational resources needed \cite{sachdeva2023data}. It involves summarizing or compressing a large dataset into a smaller, more manageable subset while retaining the most essential information needed for training models. This process aims to maintain the performance of models trained on the distilled data, ensuring that they perform similarly to models trained on the full dataset. Data distillation methods can be categorized into four main types. Meta-model matching \cite{wang2018dataset,loo2022efficient} optimizes the transferability of models trained on distilled data to the original dataset. Gradient matching \cite{zhao2021dataset,zhao2020dataset} aligns the gradients of training and distilled datasets to ensure similar model performance. Trajectory matching \cite{cui2023scaling,cazenavette2022dataset} aims to match the training trajectories of models on distilled and full datasets. Distribution matching \cite{zhao2023dataset,wang2022cafe} directly aligns the statistical distributions of the distilled and original datasets. These methods create high-fidelity, compressed datasets that retain essential information for effective machine learning model training and inference. Influence functions have not yet been used for direct dataset distillation, but they have been employed together in some similar tasks. \cite{ye2024distilled} has used a distilled dataset with a reverse gradient matching technique to approximate the computation of influence values of a smaller dataset achieving promising results. \cite{pruthi2020estimating} shows the effectiveness of TracIn methods by identifying mislabeled data and filtering them out of the dataset.

\begin{figure}[t]
\begin{center}
\hskip1ex
        \centering
    \begin{minipage}{0.34\textwidth}
        \centering
        \includegraphics[width=0.8\textwidth]{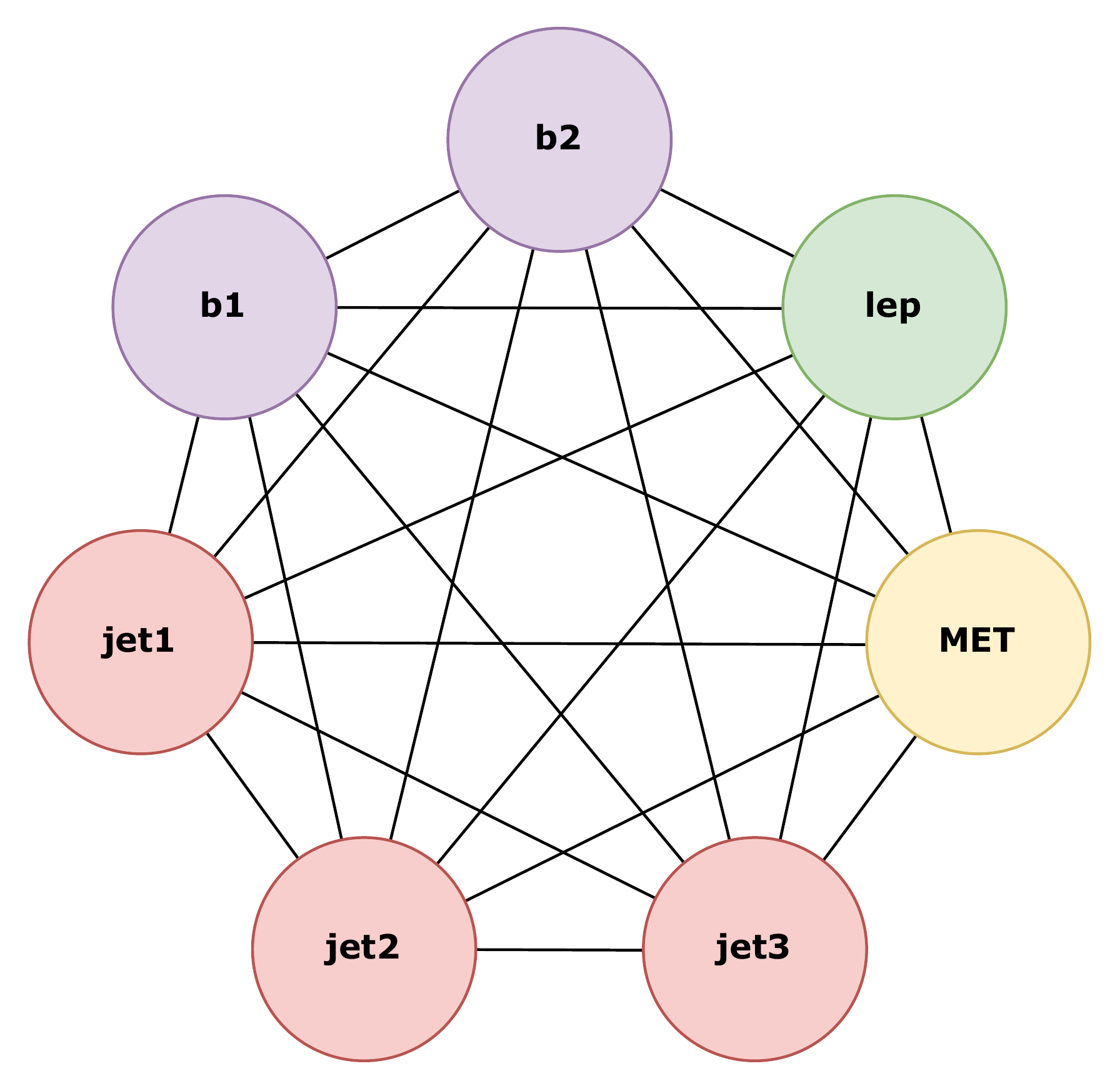}
        \caption{Complete graphs with kinematic features as nodes.}
        \label{fig:graph_event}
    \end{minipage}
    \hskip10ex
    \begin{minipage}{0.5\textwidth}
        \centering
        \begin{tabular}{lcccccc}
            \toprule
            \textbf{Particle} & \textbf{F1} & \textbf{F2} & \textbf{F3} & \textbf{F4} & \textbf{F5} & \textbf{F6} \\
            \midrule
            jet1 & $p_\mathrm{T}^\mathrm{j1}$ & $\eta^\mathrm{j1}$ & $\phi^\mathrm{j1}$ & $\mathrm{j1_{quantile}}$ & \textrm{-} & \textrm{-} \\
            jet2 & $p_\mathrm{T}^\mathrm{j2}$ & $\eta^\mathrm{j2}$ & $\phi^\mathrm{j2}$ & $\mathrm{j2_{quantile}}$ & \textrm{-} & \textrm{-} \\
            jet3 & $p_\mathrm{T}^\mathrm{j3}$ & $\eta^\mathrm{j3}$ & $\phi^\mathrm{j3}$ & $\mathrm{j3_{quantile}}$ & \textrm{-} & \textrm{-} \\
            b1   & $p_\mathrm{T}^\mathrm{b1}$ & $\eta^\mathrm{b1}$ & $\phi^\mathrm{b1}$ & $\mathrm{b1_{quantile}}$ & $\mathrm{b1_m}$ & \textrm{-} \\
            b2   & $p_\mathrm{T}^\mathrm{b2}$ & $\eta^\mathrm{b2}$ & $\phi^\mathrm{b2}$ & $\mathrm{b2_{quantile}}$ & $\mathrm{b2_m}$ & \textrm{-} \\
            lepton & $p_\mathrm{T}^\mathrm{l1}$ & $\eta^\mathrm{l1}$ & $\phi^\mathrm{l1}$ & \textrm{-} & \textrm{-} & \textrm{-} \\
            energy & $E_\mathrm{T}^\mathrm{Miss}$ & \textrm{-} & $\phi^\mathrm{ETMiss}$ & \textrm{-} & \textrm{-} & \textrm{-} \\
            \bottomrule
        \end{tabular}
        \caption{Features~\cite{ATLAS:2023act} exploited for each node.}
        \label{tab:features}
    \end{minipage}
\end{center}
\end{figure}

\section{Methodology}
\label{sec:methodology}
Our method proposes to integrate TracIn, an important data attribution method, with GNNs to enhance ATLAS analyses event classification tasks, improving performance and interpretability. We developed a three-step method: initially, we train a GNN model to classify collision event types, then we use TracIn to identify influence scores in training samples; finally, we re-train the model on a selected subset. Training elements that don't positively contribute to the classification task are then removed, improving classification metrics and reducing computational costs.

\subsection{Problem formulation}
The `SUSY dataset'~\cite {ATLAS:2023act} contains Monte Carlo simulated collision events recorded with the ATLAS experiment, representing signals over a large background with observable kinematic features. 
Two types of events are considered:
\begin{itemize}
    \item $Signal$: SuSy Dark Matter Monte Carlo candidate events
    \item $Background$: SM backgrounds form single top and top-antitop processes.
\end{itemize}
An example of signal event is presented in Fig.~\ref{fig:event_collision2d}. 
The main task involves recognizing rare signals over large backgrounds from the Standard Model processes. To recognize them, we have kinematic features that offer discriminating power in solving the task.
The particle collision event can be represented as a graph $G$: the GNN takes $G$ as input and outputs its probabilities $Y$ over classes 0, background, or 1, signal. The collision events can be represented as fully connected graphs with 6/7 nodes, $N$, and a maximum of 6 features, i.e., $X\in \mathrm{R}^{N\times 6}$. 
The particle features~\cite{ATLAS:2023act} introduced are: the transverse momentum $p_{T}^{i}$, the angular variables $\varphi^{i}$ and $\eta^{i}$, the missing transverse momentum $E_{T}^{miss}$, the mass $bi_{m}$, and the jet flavor probability $j_{quantile}$. 
By defining the graph as fully connected, we can define its adjacency matrix $A \in \mathrm{R}^{N\times N}$. Then, $G$ can be alternatively expressed as $G = (A,X)$. A graph representation and a table representing the features employed for each node is presented in Fig.~\ref{fig:graph_event} and Tab.~\ref{tab:features}.

\subsection{Framework's workflow}
\paragraph{Preliminary training.}
The GNN baseline for our experiments consists of a sequence of 2-layer Graph Convolutional operators ($GConv$) \cite{kipf2017semisupervised}, a global mean pooling operator, $GlobMeanPool$ and a final linear layer $Lin$; we used ReLU as non-linear activation. Formally, the model can be expressed as:
\begin{equation}
    \hat{y} = Lin(GlobMeanPool(GConv_2(GConv_1(X,A),A)))
\end{equation}
Once defined the model, the first step of our workflow consists of training it on the original full training set or a randomly selected subset of it: we call these approaches GNN-FT and GNN-RST respectively. This step is essential since it allows us to collect training checkpoints that are later used by the TracIn method to generate influence scores. Moreover, the experimental results obtained from both GNN-FT and GNN-RST will serve as metrics of comparison with our method.

\paragraph{Influence-based training.}
We employ the TracIn \cite{pruthi2020estimating} method as a baseline for estimating training data influence scores. It assigns an influence score to each training sample to determine its impact on the dataset. It generates influence scores via a scalable and efficient implementation: a first-order gradient approximation is performed to the exact computation of the influence values to reduce the computational cost, it utilizes checkpoints to more efficiently reproduce the training process, and finally, we choose the final layer for computing the loss gradients, i.e., the last linear layer. All these characteristics make TracIn an optimal candidate for data-intensive scenarios. For each training sample, we compute the influence score of it for itself: these values take the name of Self-influence (SI). By representing the loss function $l$, having $k$ checkpoints available, learning rate $\eta$, trainable weights $w_j$ of the $j$-th layer and training sample $x$, the Self-influence score can be computed as follow:

\begin{equation}
     SelfInfluence(x) = \sum_{i=1}^k \eta_i {\nabla l(w_{j_i},x)} {\nabla l(w_{j_i},x)} 
\end{equation}

It traces how a training point influences its own prediction: high values of self-influence scores correspond to the most diverging samples, potential outliers, mislabeled data, or more general samples with contrasting behavior. Self-influence scores have been used previously for finding mislabeled or confounding images and unsupervised anomaly detection tasks \cite{pruthi2020estimating,thimonier2022tracinad}. The main idea for our method is that by removing harmful, superfluous, or counterproductive samples, from the model and the task point of view, we can both increase model accuracy and computational efficiency.
Once we have a self-influence score for each training sample, we filter out the one with the highest values and the remaining will constitute the final training set. In this way, training elements that do not positively contribute to the classification task are removed and the final dataset will contribute to increase classification metrics and reducing computational costs. In the final step, we train the GNN baseline on the influence-based reduced training set: we named this approach GNN-IRST. 

\begin{figure}[t]
    \centering
    \subfloat[]{\includegraphics[width=0.43\textwidth]{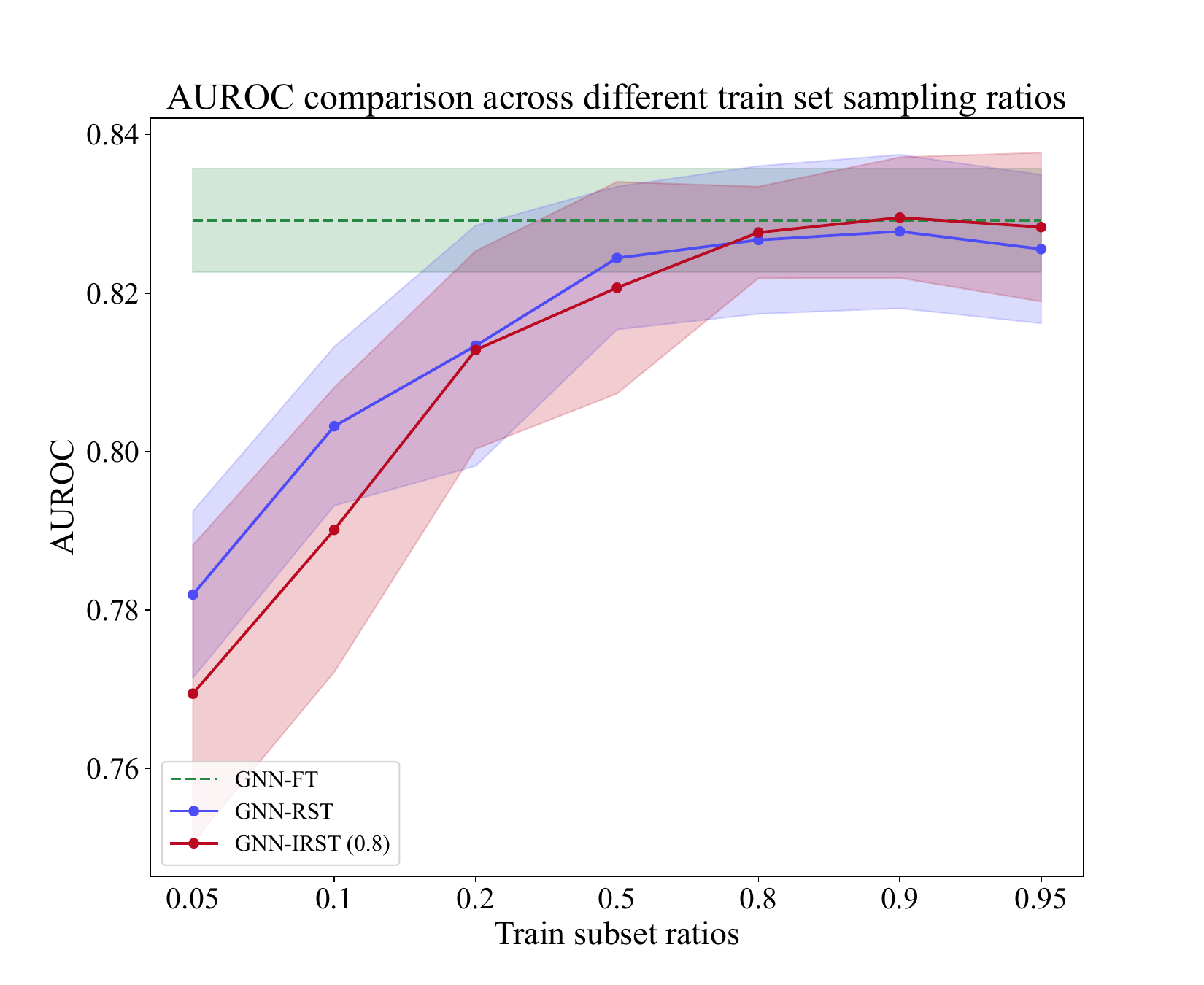}\label{fig:sub1}}\hskip1ex
    \subfloat[]{\includegraphics[width=0.43\textwidth]{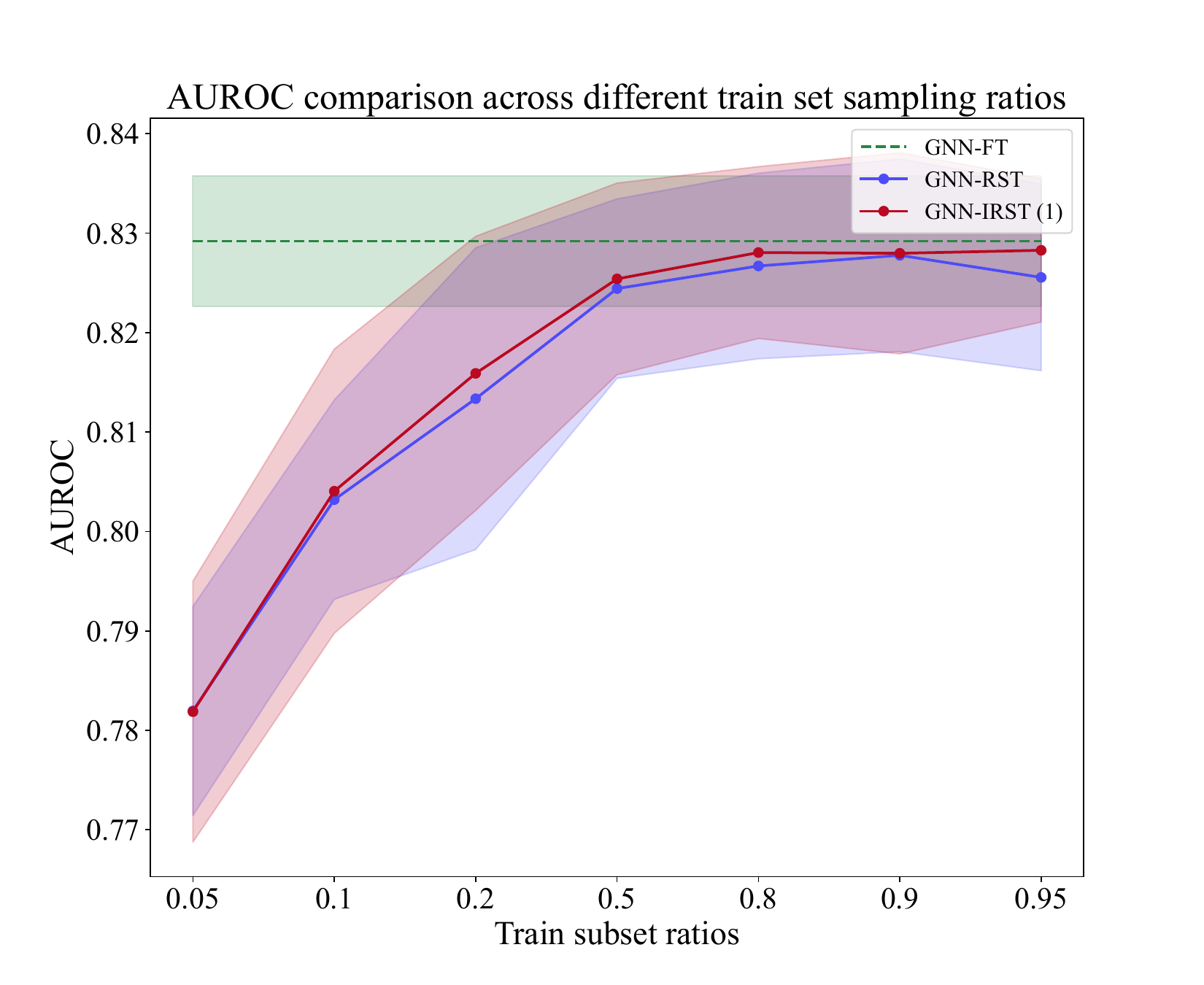}\label{fig:sub2}}
    \vspace{-6pt}
		\caption{AUROC score profile varying percentages of the initial randomly selected dataset, using 0.8 (a) and no (b) thresholds on influence values.}
		\label{fig:auroc_trend}
\end{figure}

\begin{table}[t]
    \centering
		\caption{Best performance metrics of GNN-baseline trained on full-size dataset (FT), on random-selected training subset (RST) and influence-random-selected training subset (IRST). }
    \scriptsize
        \begin{tabular}{lcccccccc}
        \toprule
        \textbf{Method} & \textbf{\%Train Samples} & \textbf{\% Influence Samples} & \textbf{\% Total Train Samples} & \textbf{Accuracy} & \textbf{F1-score} & \textbf{Precision} & \textbf{Recall} & \textbf{AUROC} \\ 
        \midrule
        GNN-FT & 100 & -- & 100 & 74.32 ± 0.7 & 72.96 ± 1.6 & 77.27 ± 3.4 & 69.65 ± 5.8 & \textbf{82.92 ± 0.7} \\ 
        GNN-RST & 80 & -- & 80 & 73.57 ± 1.2 & 72.58 ± 2.2 & 75.96 ± 4.8 & 70.47 ± 8.4 & 82.67 ± 0.9 \\ 
        GNN-IRST (Our) & 80 & 80 & 64 & 74.67 ± 1.0 & \textbf{74.18 ± 1.7} & 74.99 ± 3.7 & \textbf{74.20 ± 7.8} & 82.33 ± 0.9 \\ 
        GNN-IRST (Our) & 90 & 80 & 72 & \textbf{74.76 ± 0.7} & 73.59 ± 2.0 & \textbf{77.43 ± 3.5} & 70.72 ± 6.4 & \textbf{82.95 ± 0.8} \\ 
        \bottomrule
        \end{tabular}
    \label{table:model_comparison}
\end{table}

\section{Experimental setup}
\label{sec:experimental_setup}
We performed several experiments to validate the efficiency of our proposed methodology. In the following we will show the experimental setup, the numerical and visual results.

\subsection{Experiments' Parameters}
Our experiments included the exploration of various parameters, each contributing uniquely to the refinement of our methodology. The metrics used for the method's evaluation were Precision, Recall, F1-score, Accuracy, and AUROC. We utilized the Adam \cite{kingma2014adam} optimizer with a learning rate of $1 \cdot 10^{-3}$ and a weight decay of $5 \times 10^{-4}$. Numerical results are presented as the mean and standard deviation from 4 executions with different random seeds. The experiments ran for a maximum of 300 epochs, with 16,000 samples.

We evaluated the performance using different subset percentages of the training set: 95\%, 90\%, 80\%, 50\%, 20\%, 10\%, and 5\%. Additionally, we examined influence-based subsets with percentages of 100\%, 80\%, 50\%, and 20\%. This comprehensive analysis allowed us to assess how training set composition and influence-based selection impact the GNN model's performance, providing insights into the efficiency and effectiveness of our method. We tested three different setups:
\begin{itemize}
    \item \textbf{GNN-FT:} GNN model trained on the full train set.
    \item \textbf{GNN-RST:} GNN model trained on a random-selected subset of the original train set.
    \item \textbf{GNN-IRST} (Our): GNN model trained on the influence-based random subset of the train set.
\end{itemize}

Testing and metrics comparisons are conducted relative to the original test set size, serving as the benchmark for evaluating performance and assessing the effectiveness of various methods.

\begin{figure}[t]
    \centering
    \subfloat[]{\includegraphics[width=0.43\textwidth]{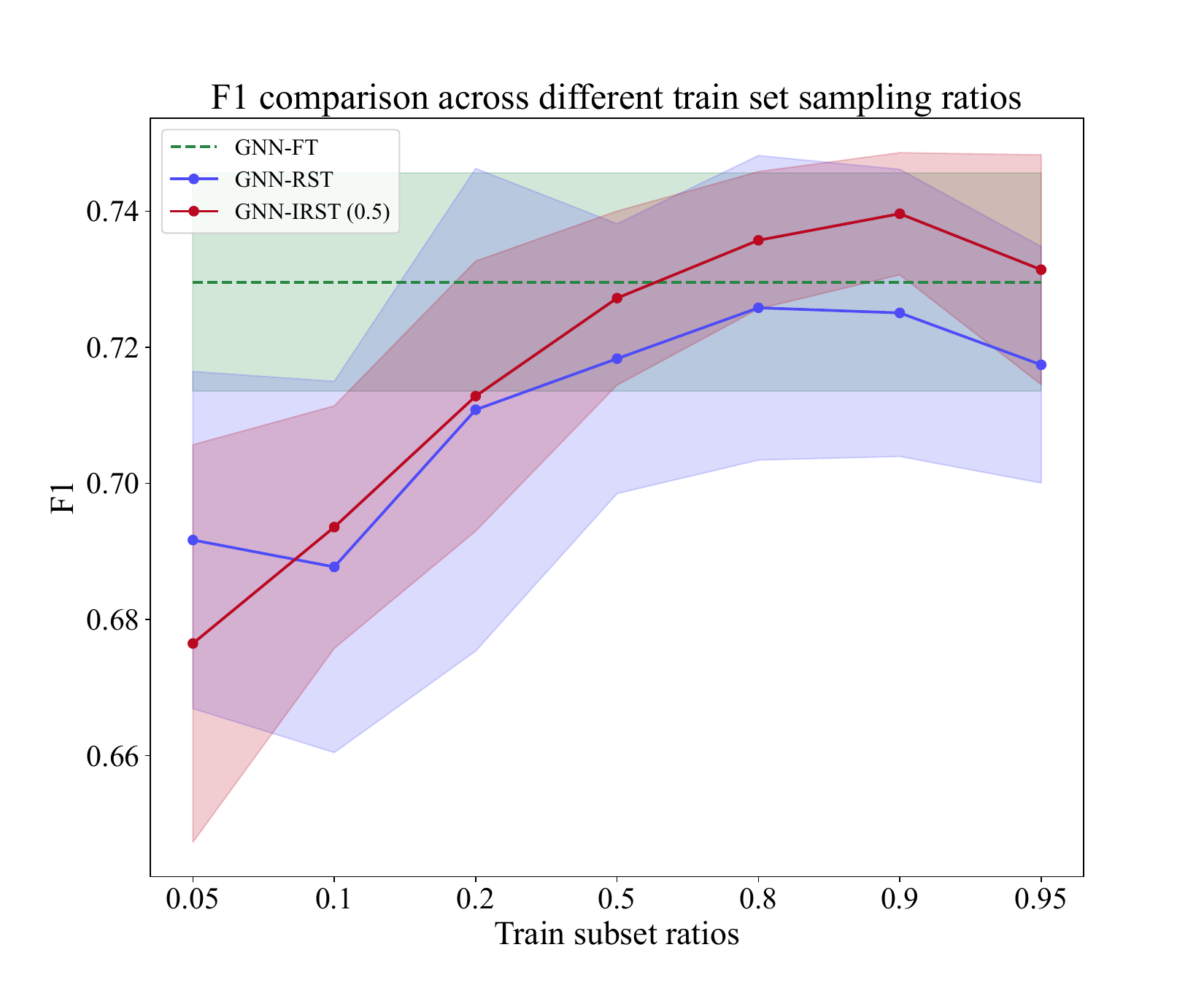}\label{fig:sub1}}\hskip1ex
    \subfloat[]{\includegraphics[width=0.43\textwidth]{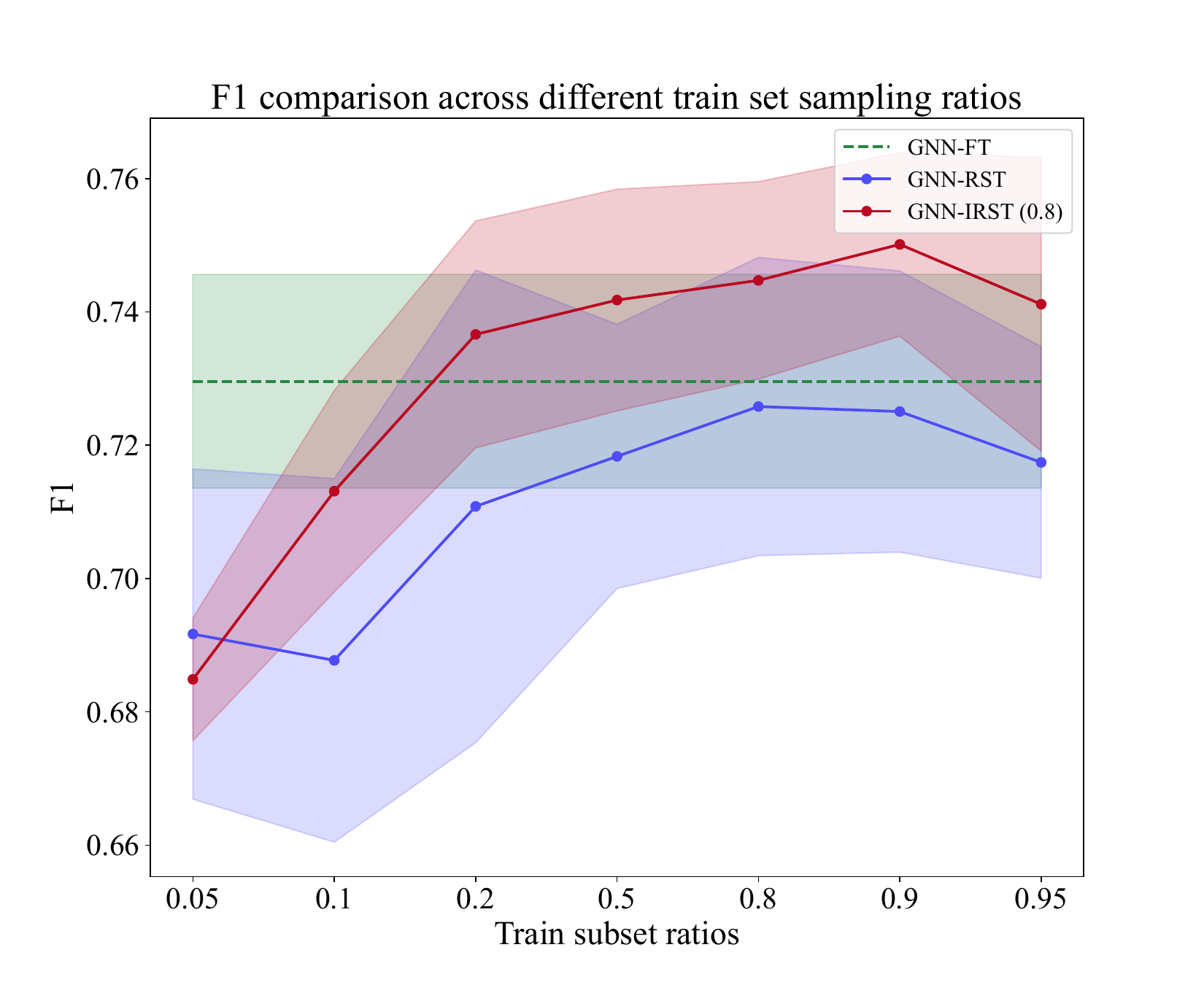}\label{fig:sub2}}
    \vspace{-6pt}
		\caption{F1-score profile varying percentages of the initial randomly selected dataset, using 0.5 (a) and 0.8 (b) thresholds on influence values.}
    \label{fig:f1_trend}
\end{figure}

\begin{table}[t]
\centering
\caption{Best performance metrics of GNN-baseline trained on different training set setups and subset percentage.}
\scriptsize
    \begin{tabular}{lcccccccc}
    \toprule
    \textbf{Method} & \textbf{\% Train sample} & \textbf{\% Influence Samples} & \textbf{\% Original trainset} & \textbf{Accuracy} & \textbf{F1-score} & \textbf{Precision} & \textbf{Recall} & \textbf{AUROC} \\ 
    \midrule
    GNN-FT & 100 & -- & 100 & \textbf{74.32 ± 0.7}  & 72.96 ± 1.6           & \textbf{77.27 ± 3.4} & 69.65 ± 5.8 & \textbf{82.92 ± 0.7} \\ 
    GNN-RST & 95 &  -- & 95   & 73.40 ± 1.5       & 71.75 ± 1.7            & 76.93 ± 3.2 & 67.40 ± 3.9 & 81.98 ± 1.0 \\ 
    GNN-IRST &    & 80 & 76      & 73.40 ± 1.1       & \textbf{73.40 ± 2.2} & 72.27 ± 2.8 & \textbf{76.72 ± 6.0} & 81.55 ± 1.6 \\ \midrule
    GNN-RST & 90 & -- & 90 & 73.60 ± 0.9          & 72.50 ± 2.1          & 76.14 ± 3.7& 69.89 ± 6.5 & 82.78 ± 1.0 \\ 
    GNN-IRST &  & 80 & 72 & \textbf{74.76 ± 0.7} & \textbf{73.59 ± 2.0} & \textbf{77.43 ± 3.5} & \textbf{70.72 ± 6.4} & \textbf{82.95 ± 0.8} \\ \midrule
    GNN-RST & 80 & -- & 80 & 73.57 ± 1.2          & 72.58 ± 2.2          & \textbf{76.00 ± 4.8} & 70.47 ± 8.4 & \textbf{82.67 ± 0.9} \\ 
    GNN-IRST &   & 80 & 64 & \textbf{74.67 ± 1.0} & \textbf{74.18 ± 1.7} & 74.99 ± 3.7          & \textbf{74.20 ± 7.8} & 82.33 ± 0.9 \\ \midrule
    GNN-RST & 50 & -- & 50 & 72.92 ± 1.1         & 71.25 ± 2.3         & \textbf{76.00 ± 2.7} & 67.37 ± 5.5 & 81.07 ± 1.6 \\ 
    GNN-IRST &   & 80 & 40 & \textbf{73.06 ± 1.0} & \textbf{74.18 ± 1.6} & 71.28 ± 0.1         & \textbf{77.57 ± 4.0} & \textbf{81.25 ± 1.7} \\ 
		\bottomrule
    \end{tabular}
\label{table:model_comparison_given_subset}
\end{table}

\subsection{Experimental results}
Primary numerical results are presented in Tab.~\ref{table:model_comparison}: we compare GNN-FT and GNN-RST results with our two best-performing models. In terms of overall accuracy on the selected metrics, our method consistently achieves good results, showcasing its robustness and reliability. By employing only 64\% of the original training set we're able to achieve a higher F1-score and Recall score with respect to GNN-FT, while with a percentage of 72\% we achieve also better accuracy, Precision and AUROC score. Moreover, this last setup is able to outperform for all the metrics the GNN-RST methodology, using also 8\% less training data. In  Tab.~\ref{table:model_comparison_given_subset} we perform a deep comparison of GNN-IRST performance with respect to GNN-FT and GNN-RST with different training subset percentages. For almost all the metrics and setups, an influence-based selection mechanism achieves better results with respect to the random one: this shows how a careful choice of training data can increase the overall performance of this method. We show some plots representing the evolution of AUROC and F1-score metric respectively at different data percentages in Fig.~\ref{fig:auroc_trend} and Fig.~\ref{fig:f1_trend} respectively. In Fig.~\ref{fig:auroc_trend} we can observe a slight improvement of the AUROC metric by employing 20\% less training sample with respect to the random-based subset selection. In Fig.~\ref{fig:f1_trend}a and Fig.~\ref{fig:f1_trend}b we can observe instead better performances for almost all subset percentage values; moreover, here we achieve better F1-score results with respect to the full-size training set setup at many different subset percentage values.

\section{Conclusions}
\label{sec:conclusions}
Utilizing data-attribution methods for data selection tasks proves to be an efficient strategy for enhancing the performance of GNN models in scenarios defined by a large amount of information. By precisely selecting a smaller subset of the original dataset based on self-influence values, our method can achieve competitive or superior results compared to approaches which use the full training set or a random reduced version of it.
Carefully selecting training data can significantly improve accuracy as well as the computational and operational costs of data-intensive problems, as proven with LHC experiments. This precise selection also aids in the effective downstream processing of vast amounts of data.
Moving forward, our efforts will focus on the development of a custom architecture able to manage and leverage the full potential of influence information or also on exploring and implementing various data influence methods: these techniques not only expand the traditional learning framework but also enhance the interpretability of the inner workings of these networks. This attribute is particularly demanded in intricate and critical scenarios like high-energy physics.

\section*{Acknowledgment}
The work is partly funded by the European Union’s CHIST-ERA programme under grant agreement CHIST-ERA-19-XAI-009 (MUCCA). 
AV is supported in part by the Italian Ministry of University and Research (MUR), which funded his PhD as per the Ministerial Decree no. 1061/2021. 
SG is supported by PNRR MUR project PE0000013-FAIR. SS is partly funded by Sapienza grants RM1221816BD028D6 (DESMOS) and RG123188B3EF6A80 (CENTS).

%

\end{document}